\documentclass[sigconf]{acmart}

\AtBeginDocument{%
  \providecommand\BibTeX{{%
    \normalfont B\kern-0.5em{\scshape i\kern-0.25em b}\kern-0.8em\TeX}}}

\setcopyright{rightsretained}
\copyrightyear{2021}
\acmYear{2021}
\acmDOI{}

\acmConference[SiKDD '21]{Ljubljana '21: Slovenian KDD Conference on Data Mining and Data Warehouses}{October, 2021}{Ljubljana, Slovenia}
\acmBooktitle{Ljubljana '21: Slovenian KDD Conference on Data Mining and Data Warehouses, October, 2021, Ljubljana, Slovenia}
\acmPrice{}
\acmISBN{}

\usepackage{listings}
\usepackage{xcolor}
\usepackage{adjustbox}
\usepackage{float}
\usepackage{color,soul}
\usepackage{aliascnt}
\usepackage{amsmath}
\usepackage{textgreek}
\usepackage{caption}
\usepackage{tikz-cd}
\usepackage{graphicx}
\usepackage{multirow}
\usepackage{hyperref}
\usepackage{academicons}
\restylefloat{table}

\begin{document}

\title{Active Learning for Automated Visual Inspection of Manufactured Products}

\author{Elena Trajkova}
\authornote{Both authors contributed equally to this research.}
\orcid{0000-0001-5342-1085}
\affiliation{%
  \institution{University of Ljubljana, Faculty of Electrical Engineering}
  \streetaddress{Tr\v{z}a\v{s}ka 25}
  \city{Ljubljana}
  \country{Slovenia}
  \postcode{1000}
}
\email{trajkova.elena.00@gmail.com}

\author{Jo\v{z}e M. Ro\v{z}anec}
\authornotemark[1]
\orcid{0000-0002-3665-639X}
\affiliation{%
  \institution{Jo\v{z}ef Stefan International Postgraduate School}
  \streetaddress{Jamova 39}
  \city{Ljubljana}
  \country{Slovenia}}
\email{joze.rozanec@ijs.si}

\author{Paulien Dam}
\orcid{0000-0001-5378-8100}
\affiliation{%
  \institution{Philips Consumer Lifestyle BV}
  \streetaddress{Oliemolenstraat 5}
  \city{Drachten}
  \country{The Netherlands}}
\email{paulien.dam@philips.com}

\author{Bla\v{z} Fortuna}
\orcid{0000-0002-8585-9388}
\affiliation{%
  \institution{Qlector d.o.o.}
  \streetaddress{Rov\v{s}nikova 7}
  \city{Ljubljana}
  \country{Slovenia}}
\email{blaz.fortuna@qlector.com}

\author{Dunja Mladeni\'{c}}
\orcid{0000-0003-4480-082X}
\affiliation{%
  \institution{Jo\v{z}ef Stefan Institute}
  \streetaddress{Jamova 39}
  \city{Ljubljana}
  \country{Slovenia}}
\email{dunja.mladenic@ijs.si}

\renewcommand{\shortauthors}{Trajkova and Ro\v{z}anec}

\begin{abstract}
Quality control is a key activity performed by manufacturing enterprises to ensure products meet quality standards and avoid potential damage to the brand's reputation. The decreased cost of sensors and connectivity enabled an increasing digitalization of manufacturing. In addition, artificial intelligence enables higher degrees of automation, reducing overall costs and time required for defect inspection. In this research, we compare three active learning approaches and five machine learning algorithms applied to visual defect inspection with real-world data provided by \textit{Philips Consumer Lifestyle BV}. Our results show that active learning reduces the data labeling effort without detriment to the models' performance.
\end{abstract}




\maketitle



\section{Introduction}
Quality control is one of the critical activities that must be performed by manufacturing enterprises \cite{yang2020using,wuest2014approach}. The main purpose of such activity is to detect product defects meeting quality standards, avoid rework, supply chain disruptions, and avoid potential damage to the brand's reputation \cite{benbarrad2021intelligent,wuest2014approach}. Along with the information regarding defective products, it provides insights into when and where such defects occur, which can be used to further dig into the root causes of such defects and mitigation actions to improve the quality of manufacturing products and processes.

The decreased cost of sensors and connectivity enabled an increasing digitalization of manufacturing \cite{benbarrad2021intelligent}, which along with the adoption of Artificial Intelligence (AI) \cite{grangel:2019}, represents an opportunity towards enhancing the defect detection in industrial settings \cite{chouchene2020artificial}. While the quality of the manual inspection has low scalability (requires time to train an inspector, the employees can work a limited amount of time and are subject to fatigue, and the inspection itself is slow), its quality can be affected by the operator-to-operator inconsistency, and it depends on the complexity of the task, the employees (e.g., their intelligence, experience, well-being), the environment (e.g., noise and temperature), the management support and communication \cite{see2012visual}; none of these factors affect the outcome of automated quality inspection. Machine learning has been successfully applied to defect detection in a wide range of scenarios \cite{ravikumar2011machine,duan2012machine,gobert2018application,iglesias2018automated,beltran2020external}.

An annotated dataset must be acquired to implement machine learning models for defect detection successfully. The increasing number of sensors provides large amounts of data. As the manufacturing process quality increases, the data obtained from the sensors is expected to be highly imbalanced: most of the data instances will correspond to non-defective products, and a small proportion of them will correspond to different kinds of defects. Annotating all the data is prone to similar limitations as manual inspection described in the paragraph above. It is thus imperative to provide strategies to select a limited subset of them that are most informative to the defect detection models.

We frame the defect detection problem as a supervised learning problem. Given a large amount of unlabeled data, and based on the premise that only a tiny fraction of the data provides new information to the model and thus has the potential to enhance its performance, we adopt an active learning approach. Active learning is a subfield of machine learning that attempts to identify the most informative unlabeled data instances, for which labels are requested some \textit{oracle} (e.g., a human expert) \cite{settles2009active}. This research compares three active learning strategies: pool-based sampling, stream-based sampling, and query by committee.

The main contributions of this research are (i) a comparative study between the five most frequently cited machine learning algorithms for automated defect detection and (ii) three active learning approaches (iii) for a real-world multiclass classification problem. We develop the machine learning models with images provided by the \textit{Philips Consumer Lifestyle BV} corporation. The dataset comprises shaver images divided into three classes, based on the defects related to the printing of the logo of the \textit{Philips Consumer Lifestyle BV} corporation: good shavers, shavers with double printing, and shavers with interrupted printing.

We evaluate the models using the area under the receiver operating characteristic curve (AUC ROC, see \cite{BRADLEY19971145}). AUC ROC is widely adopted as a classification metric, having many desirable properties such as being threshold independent and invariant to a priori class probabilities. We measure AUC ROC considering prediction scores cut at a threshold of 0.5.

This paper is organized as follows. Section 2 outlines the current state of the art and related works, Section 3 describes the use case, and Section 4 provides a detailed description of the methodology and experiments. Finally, section 5 outlines the results obtained, while Section 6 concludes and describes future work.

\section{Related Work}
Among the many techniques used for automated defect inspection, we find the automated visual inspection, which refers to image processing techniques for quality control, usually applied in the production line of manufacturing industries \cite{beltran2020external}. Visual inspection requires extracting features from the images, which are used to train the machine learning model. This procedure is simplified when using deep learning models, enabling end-to-end learning, where a single architecture can perform feature extraction and classification \cite{long2015fully,glasmachers2017limits}.

The use of automated visual inspection for defect detection has been applied to multiple manufacturing use cases. \cite{ravikumar2011machine} manually extracted features (e.g., histograms) from machine component images and compared the performance of the N\"aive Bayes and C4.5 models. \cite{duan2012machine} extracted statistical features from the images and compared the performance of Support Vector Machines (SVM), Multilayer Perceptron (MLP), and k-nearest neighbors (kNN) models for visual inspection of microdrill bits in printed circuit board production. \cite{gobert2018application} used 3D convolutional filters applied on computed tomography images and an SVM classifier for defect detection during metallic powder bed fusion in additive manufacturing. \cite{iglesias2018automated} used some heuristics to detect regions of interest on slate slab images, on which they performed feature engineering to later train an SVM model on them. Finally, \cite{beltran2020external} reported using a custom neural network for feature extraction and an SVM model for classification when inspecting aerospace components.

While the authors cited above worked with fully labeled datasets, a production line continually generates new data, exceeding the labeling capacity. A possible solution to this issue is the use of active learning, where the active learner identifies informative unlabeled instances and requests labels to some \textit{oracle}. Typical scenarios involve (i) membership query synthesis (a synthetic data instance is generated), (ii) stream-based selective sampling (the unlabeled instances are drawn one at a time, and a decision is made whether a label is requested, or the sample is discarded), and (iii) pool-based selective sampling (queries samples from a pool of unlabeled data). Among the frequently used querying strategies, we find (i) uncertainty sampling (select an unlabeled sample with the highest uncertainty, given a certain metric or machine-learning model\cite{lewis1994heterogeneous}), or (ii) query-by-committee (retrieve the unlabeled sample with the highest disagreement between a set of forecasting models (\textit{committee})) \cite{cohn1994improving,settles2009active}.

Active learning was successfully applied in the manufacturing domain, but scientific literature remains scarce on this domain \cite{meng2020machine}. Some use cases include the automatic optical inspection of printed circuit boards\cite{dai2018towards} and the identification of the local displacement between two layers on a chip in the semi-conductor industry\cite{van2018active}.

The use of machine learning automates the defect detection, and active learning enables an \textit{inspection by exception} \cite{chouchene2020artificial}, only querying for labels of the images that the model is most uncertain about. While this considerably reduces the volume of required inspections, it is also essential to consider that it can produce an incomplete ground truth by missing the annotations of defective parts classified as false negatives and not queried by the active learning strategy \cite{cordier2021active}.

\section{Use Case}


The use case provided for this research corresponds to visual inspection of shavers produced by \textit{Philips Consumer Lifestyle BV}. The visual quality inspection aims to detect defective printing of a logo on the shavers. This use case focuses on four pad printing machines setup for a range of different products, and different logos. A lot of products are produced every day on these machines, which are manually handled and inspected on their visual quality and removed from further processing if the prints on the products are not classified as good.  Operators spend several seconds handling, inspecting, and labeling the products. Given an automated visual quality inspection system would strongly reduce the need to manually inspect and label the images, it could speed up the process for more than 40\%. Currently there are two types of defects classified related to the printing quality of the logo on the shaver: double printing, and interrupted printing. Therefore, images are classified into three classes: good printing (class zero), double printing (class one), and interrupted printing (class two). A labeled dataset with a total of 3.518 images was provided to train and test the models.


\section{Methodology}\label{S:METHODOLOGY}
We pose automated defect detection as a multiclass classification problem. We measure the model's performance with the AUC ROC metric, using the "one-vs-rest" heuristic method, which involves splitting the multiclass dataset into multiple binary classification problems. Furthermore, we calculate the metrics for each class and compute their average, weighted by the number of true instances for each class.

To extract features from the images, we make use of the ResNet-18 model \cite{he2016deep}, extracting embeddings from the Average Pooling layer. Since the embedding results in 512 features, which could cause overfitting, we use the mutual information to evaluate the most relevant ones and select the \textit{top K} features, with $K=\sqrt{N}$, where N is the number of data instances in the train set, as suggested in \cite{hua2005optimal}. 

To evaluate the models' performance across different active learning strategies, we apply a stratified k-fold cross validation \cite{zeng2000distribution}, using one fold for testing, one fold as a pool of unlabeled data for active learning, and the rest from training the model. We adopt \textit{k=10} based on recommendations by \cite{kuhn2013applied}, and query all available unlabeled instances to evaluate the active learning approaches. We compare three active learning scenarios: drawing queries through (i) stream-based classifier uncertainty sampling accepting instances with an uncertainty threshold above the 75\textsuperscript{th} percentile of observed instances, (ii) pool-based sampling selecting the instances a given model is most uncertain about, and pool-based sampling considering a query-by-committee strategy, where the committee is created with models trained with the five algorithms we consider in this research: Gaussian N\"aive Bayes, CART (\textit{Classification and Regression Trees}, similar to C4.5, but it does not compute rule sets), Linear SVM, MLP, and kNN. Finally, we compare the performance of the active learning scenarios computing the average AUC ROC of each fold, and assess if the results differences obtained from each model are statistically significant by using the Wilcoxon signed-rank test\cite{wilcoxon1992individual}, using a p-value of 0.05.

\section{Results and Analysis}

\begin{figure*}[!ht]
\centering
\includegraphics[width=0.9\textwidth,height=0.26\textwidth]{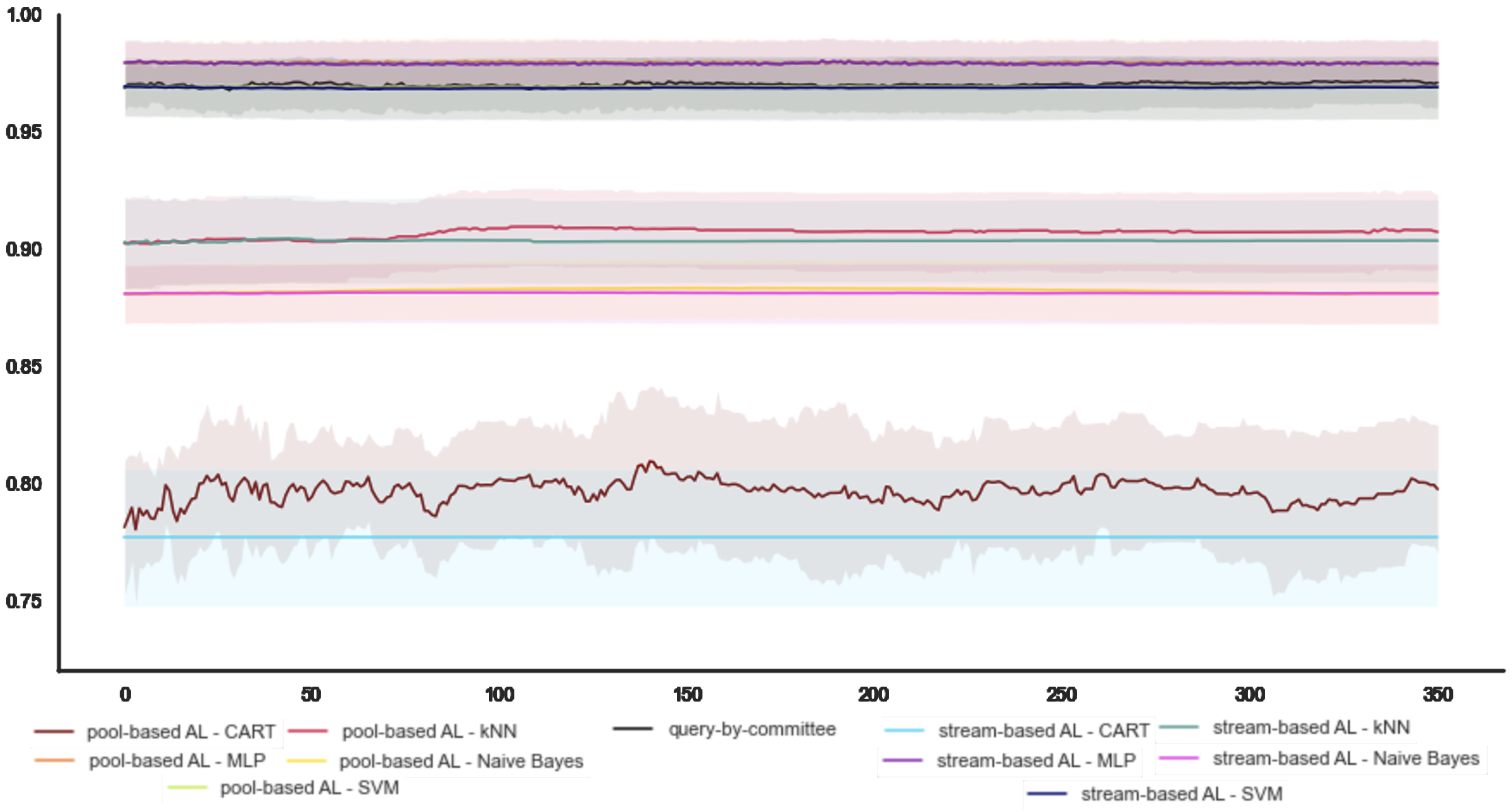}
\caption{AUC ROC mean and variance over time across the active learning scenarios and models measured for the test fold when consuming the 350 unlabeled samples.}
\label{F:RESULTS-AUC}
\end{figure*} 

\begin{table*}[ht!]
\resizebox{0.9\textwidth}{!}{
\begin{tabular}{|l|l|r|r|r|r|r|r|r|r|r|r|}
\hline
\textbf{Active Learning scenario} & \textbf{Model} & \multicolumn{1}{l|}{\textbf{Fold 1}} & \multicolumn{1}{l|}{\textbf{Fold 2}} & \multicolumn{1}{l|}{\textbf{Fold 3}} & \multicolumn{1}{l|}{\textbf{Fold 4}} & \multicolumn{1}{l|}{\textbf{Fold 5}} & \multicolumn{1}{l|}{\textbf{Fold 6}} & \multicolumn{1}{l|}{\textbf{Fold 7}} & \multicolumn{1}{l|}{\textbf{Fold 8}} & \multicolumn{1}{l|}{\textbf{Fold 9}} & \multicolumn{1}{l|}{\textbf{Fold 10}} \\ \hline
\multirow{5}{*}{\textbf{stream-based}} & \textbf{CART} & 0,8168 & 0,7828 & 0,7810 & 0,7694 & 0,8196 & 0,7805 & 0,7843 & 0,7970 & 0,8409 & 0,7940 \\ \cline{2-12} 
 & \textbf{kNN} & 0,9289 & 0,9121 & 0,9174 & 0,8686 & 0,9024 & 0,9000 & 0,9051 & 0,8960 & 0,9282 & 0,9082 \\ \cline{2-12} 
 & \textbf{MLP} & \textbf{0,9900} & \textbf{0,9928} & \textbf{0,9846} & \textbf{0,9563} & \textbf{0,9804} & \textbf{0,9807} & \textbf{0,9710} & \textbf{0,9729} & 0,9793 & \textbf{0,9845} \\ \cline{2-12} 
 & \textbf{N\"aive Bayes} & 0,8818 & 0,8668 & 0,8819 & 0,8686 & 0,8829 & 0,8899 & 0,8650 & 0,8877 & 0,8864 & 0,9098 \\ \cline{2-12} 
 & \textbf{SVM} & 0,9752 & 0,9828 & 0,9725 & \textit{0,9530} & 0,9816 & 0,9720 & 0,9570 & 0,9412 & 0,9824 & 0,9712 \\ \hline
\multirow{5}{*}{\textbf{pool-based}} & \textbf{CART} & 0,7584 & 0,7904 & 0,7543 & 0,7468 & 0,8441 & 0,7730 & 0,8044 & 0,7701 & 0,7850 & 0,7412 \\ \cline{2-12} 
 & \textbf{kNN} & 0,9189 & 0,9149 & 0,9161 & 0,8581 & 0,9055 & 0,9036 & 0,8961 & 0,8910 & 0,9224 & 0,9056 \\ \cline{2-12} 
 & \textbf{MLP} & \textit{0,9892} & \textit{0,9921} & \textit{0,9845} & \textbf{0,9563} & 0,9790 & \textbf{0,9803} & \textbf{0,9702} & \textbf{0,9723} & 0,9806 & \textit{0,9840} \\ \cline{2-12} 
 & \textbf{N\"aive Bayes} & 0,8800 & 0,8654 & 0,8809 & 0,8677 & 0,8813 & 0,8895 & 0,8637 & 0,8873 & 0,8850 & 0,9090 \\ \cline{2-12} 
 & \textbf{SVM} & 0,9752 & 0,9819 & 0,9726 & 0,9518 & \textit{0,9806} & 0,9712 & 0,9562 & 0,9412 & \textit{0,9823} & 0,9722 \\ \hline
\multicolumn{2}{|l|}{\textbf{query-by-committee}} & 0,9774 & 0,9824 & 0,9714 & 0,9500 & 0,9723 & \textit{0,9726} & \textit{0,9597} & \textit{0,9571} & \textbf{0,9830} & 0,9734 \\ \hline
\end{tabular}
\caption{AUC ROC values were obtained across the ten cross-validation folds. Best results are bolded, second-best results are highlighted in italics. \label{T:AUC-ROC-VALUES}}}
\end{table*}

\begin{table*}[ht!]
\resizebox{0.9\textwidth}{!}{
\begin{tabular}{|l|r|r|r|}
\hline
\multicolumn{1}{|c|}{\multirow{2}{*}{\textbf{Model}}} & \multicolumn{3}{l|}{\textbf{Active Learning scenarios}} \\ \cline{2-4} 
\multicolumn{1}{|c|}{} & \multicolumn{1}{l|}{\textbf{stream-based vs. pool-based}} & \multicolumn{1}{l|}{\textbf{stream-based vs. query-by-committee}} & \multicolumn{1}{l|}{\textbf{pool-based vs. query-by-committee}} \\ \hline
\textbf{CART} & 0,0840 & \textbf{0,0020} & \textbf{0,0020} \\ \hline
\textbf{kNN} & 0,1309 & \textbf{0,0020} & \textbf{0,0020} \\ \hline
\textbf{MLP} & 0,0856 & \textbf{0,0039} & \textbf{0,0039} \\ \hline
\textbf{N\"aive Bayes} & \textbf{0,0020} & \textbf{0,0020} & \textbf{0,0020} \\ \hline
\textbf{SVM} & 0,1824 & 0,4316 & 0,6250 \\ \hline
\end{tabular}
\caption{p-values obtained for the Wilcoxon signed-rank test when comparing the average of AUC ROC results across ten cross-validation folds. \label{T:P-VALUES}}}
\end{table*}

The results obtained from the experiments we ran, and described in Section \ref{S:METHODOLOGY}, are presented in Table \ref{T:AUC-ROC-VALUES}, Table \ref{T:P-VALUES}, and Fig. \ref{F:RESULTS-AUC}. Table \ref{T:AUC-ROC-VALUES} describes the average AUC ROC per each active learning scenario and model for each cross-validation test fold. We observe that the best model across strategies is the MLP, which achieved the best or second-best performance across almost every fold in pool-based and stream-based active learning. Among those two scenarios, the best results were obtained for stream-based active learning. We observed the same across the rest of the models, though the differences were not significant for all but the N\"aive Bayes models (see Table \ref{T:P-VALUES}). Query-by-committee displayed a strong performance, showing best results immediately after the MLP. When assessing the statistical significance between the query-by-committee scenario and results obtained from different models with stream-based and pool-based strategies, we observed that differences were significant in all cases, except for the SVM models. SVM models, most widely used in active learning literature related to automated defect inspection, were the third-best models among the tested ones, immediately after the MLPs in stream-based and pool-based active learning and the query-by-committee approach. SVM models did not display significant differences when compared across different active learning scenarios. The worst results were consistently observed for the CART models.

When analyzing the results, we were interested in how the models' performance evolved through time and significant variations between the first and last results observed. To that end, we assessed the statistical significance between the means of the first and last quartiles of the test fold for each active learning scenario. We assessed the statistical significance using the Wilcoxon signed-rank test, with a p-value of 0.05. While such variations existed and were positive in most test folds (the models learned through time), the improvements were not statistically significant in none of the scenarios.

\section{Conclusion}
In this paper, we compared three active learning scenarios (pool-based, stream-based with classifier uncertainty sampling, and query-by-committee) across five machine learning algorithms (Gaussian N\"aive Bayes, CART, Linear SVM, MLP, and kNN). We found that the best performance was achieved by the MLP model regardless of the active learning strategy. The second-best performance was obtained through the query-by-committee strategy, while the frequently used SVM models ranked third. We found no significant difference between using pool-based or stream-based active learning approaches. Results from the query-by-committee approach were statistically significant in all cases and better than all the models, except for the MLPs. Finally, we found no case where the improvement between the first and last quartile of the test fold in each active learning scenario would be significant. Future work will focus on data augmentation techniques that could help achieve a statistically significant improvement over time when applying active learning techniques.

\begin{acks}
This work was supported by the Slovenian Research Agency and the European Union’s Horizon 2020 program project STAR under grant agreement number H2020-956573. The authors acknowledge the valuable input and help of Jelle Keizer and Yvo van Vegten from \textit{Philips Consumer Lifestyle BV}.
\end{acks}

\bibliographystyle{ACM-Reference-Format}
\bibliography{main}

\end{document}